\documentclass{article}
\usepackage{geometry}
\usepackage[pdftex]{graphicx}
\usepackage{subcaption}
\usepackage{cite}
\usepackage{array}

\begin{document}

\title{Bayesian Unification of Gradient and Bandit-based Learning for Accelerated Global Optimisation}

%\author{\IEEEauthorblockN{Ole-Christoffer Granmo}
%\IEEEauthorblockA{Department of ICT\\
%University of Agder\\
%Norway\\
%Email: ole.granmo@uia.no}}

\date{}

\author{
Ole-Christoffer Granmo\thanks{Author's status: {\it Professor}. This author can be contacted at: Centre for Artificial Intelligence Research (CAIR), University of Agder, Grimstad, Norway.  E-mail: {\tt ole.granmo@uia.no}.}}

\maketitle

\begin{abstract}
Bandit based optimisation schemes have a remarkable advantage over gradient based approaches due to their global perspective, which eliminates the danger of getting stuck at local optima. However, for continuous optimisation problems or problems with a large number of 
actions, bandit based approaches can be hindered by slow learning. 
Gradient based approaches, on the other hand, navigate quickly in high-dimensional continuous spaces through local optimisation, following the gradient in fine grained steps. However, apart from being susceptible to local optima, these schemes are also less suited for online learning due to their reliance on extensive trial-and-error before the optimum can be identified. In contrast, bandit algorithms seek to identify the optimal action (global optima) in as few steps as possible.  In this paper, we propose a Bayesian approach that unifies the above two distinct paradigms in one single framework, with the aim of combining their advantages. At the heart of our approach we find a stochastic linear approximation of the function to be optimised, where both the gradient and values of the function are explicitly captured.  This model allows us to learn from both noisy function and gradient observations, as well as predicting these properties across the action space to support optimisation. We further propose an accompanying bandit driven exploration scheme that uses Bayesian credible bounds to trade off exploration against exploitation.  Our empirical
results demonstrate that by unifying bandit and gradient based learning, one obtains consistently improved performance across a wide spectrum of problem environments. Furthermore, even when gradient feedback is unavailable, the flexibility of our model, including gradient prediction, still allows us outperform competing approaches, although with a smaller margin. Due to the pervasiveness of bandit based optimisation, our scheme  opens up for improved performance both in meta-optimisation and in applications where gradient related information is readily available.
\end{abstract}


% \fi
%\IEEEpeerreviewmaketitle

\section{Introduction}

\subsection{Background and Motivation}
The multi-armed bandit problem is a classical optimisation problem that captures the trade off between exploitation and exploration in reinforcement learning. The problem consists of an agent that sequentially pulls one out of multiple arms attached to a gambling machine, with each pull resulting in a scalar reward. Each reward is randomly drawn from an \emph{unknown} distribution, unique to each arm. The purpose is to as quickly as possible identify the arm with the highest expected reward, through goal directed trial-and-error.

Bandit based optimisation schemes have a tremendous advantage over gradient based approaches (such as \cite{bengio2000}) due to their global perspective, which eliminates the danger of getting stuck at local optima. However, for continuous optimisation problems or problems with a large number of arms (actions), bandit based approaches are hindered by their inability to generalise across arms (typically modelling arms as independent reward sources). This independence assumption leads to slow learning because the expected reward function must be inferred independently for each arm. Gradient based approaches, on the other hand, navigate quickly in high-dimensional continuous spaces through {\it local} optimisation, following the gradient in small steps. The local optimisation, however, makes this class of schemes susceptible to local optima. Further, they are less suited for on-line learning due to their reliance on extensive trial and error, with small parameter adjustments at each step. In contrast, bandit algorithms are designed for on-line operation, aiming to converge to the optimal arm (global optima) in as few trials as possible.

To deal with continuous and large action spaces, several bandit based approaches have recently been proposed that capture interaction among actions. One class of schemes, referred to as \emph{global} multi-armed bandit schemes, models the expected rewards of the arms as (non-)linear functions of a global parameter $\gamma$ \cite{Atan2015}. Another family of techniques attacks large action spaces through tree based searching, with X-Armed Bandits finding global maxima when the expected reward (objective) function is "locally Lipschitz" \cite{Bubeck2011}. Finally,  Gaussian processes have been applied for smoothing and interpolation, forming the foundation for bandit based exploration and exploitation in continuous action spaces \cite{Srinivas2012}.

\begin{figure}[!t]
\centering
\includegraphics[width=4.0in]{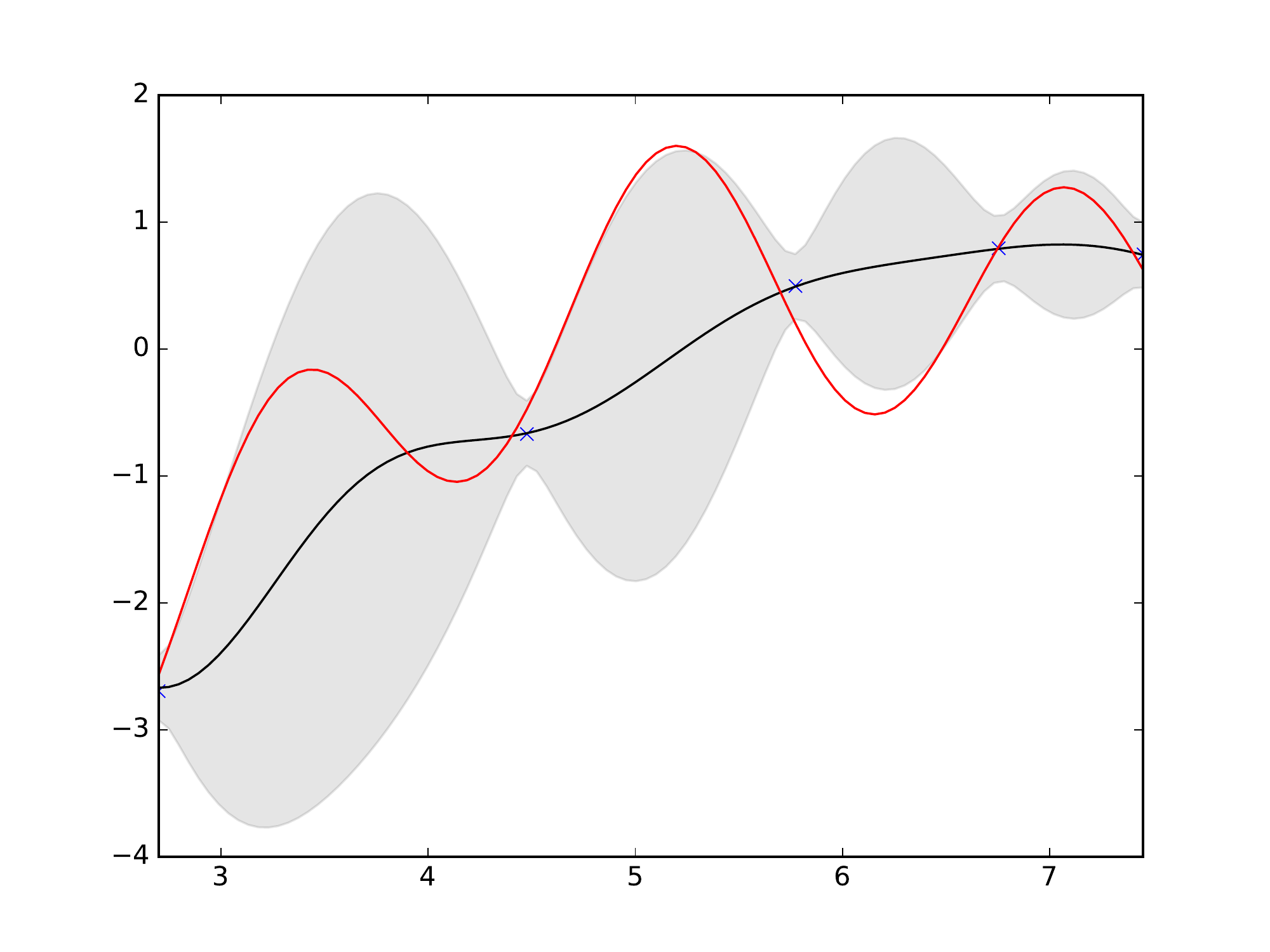}
\caption{Gaussian process with a squared exponential covariance function (isotropic distance and white noise). The red line marks the original function, while the black line marks the function predicted from five noisy observations (blue crosses) along with a 99\% credible interval.}
\label{fig_gp}
\end{figure}

Gaussian process based approaches are particularly attractive because they provide a Bayesian estimate of the expected reward functions including credible intervals, as illustrated in Fig. \ref{fig_gp}.
In brief, dedicated kernel functions capture smoothness and other function dynamics, encoded in a covariance matrix.  However, as illustrated in the figure,  the scheme is "blind" towards the gradient of the underlying reward function (red line), merely "tracing" a line through the observations (crosses),  tending towards a prior mean without other input (typically set to zero). 

In conclusion, gradient and bandit based scheme have distinct advantages and disadvantages.

\subsection{Paper Contributions and Outline}

In this paper  we propose a radically new approach to stochastic optimisation where the global perspective of multi-armed bandits is combined with gradient based local optimisation, with the effect of significantly accelerating learning. In all brevity, our approach provides a Bayesian Unification of Gradient and Bandit-based learning (hereafter referred to as BUG-B). Our contributions can be summarised as follows:
\begin{itemize}
\item At the heart of BUG-B we find a novel Bayesian model that explicitly connects the expected reward function with its gradient. The model supports learning from both noisy function values as well as gradient related observations. Further, unobserved function values and gradient information can be predicted across the action space to support goal-directed exploration and exploitation of the reward function.
\item In addition, we propose an accompanying bandit driven exploration scheme that uses Bayesian credible bounds to trade off exploration against exploitation. Note that BUG-B also lends itself towards so-called Thompson Sampling  \cite{Granmo2010a,Scott2010,May2011,Thompson1933}
 due to its Bayesian nature.
\item Our empirical
results demonstrate that by unifying bandit and gradient based learning, one obtains improved performance across a wide spectrum of reward functions and degrees of noise.
\item Even when gradient feedback is unavailable, the flexibility of our model, including gradient estimation, allows us to still outperform competing approaches, although with a smaller margin.
\end{itemize}

In Section \ref{BUG-B} we provide the details of our Bayesian approach to unifying gradient and bandit-based learning. We introduce a grid based linear approximation of the reward function that explicitly relates function- and gradient values, modelled as a set of stochastic variables to address noisy observations and relationships. We then cover accompanying optimisation strategies based on Bayesian credibility bounds as well as Thompson Sampling, before we in Section \ref{results} provide empirical results demonstrating the superiority of our scheme in a wide range of settings. We conclude the paper in Section \ref{conclusion} by providing pointers for further research.

\section{Bayesian Unification of Gradient and Bandit-based Learning (BUG-B)}
\label{BUG-B}

\subsection{The BUG-B Model}

The BUG-B model is based on a linear approximation of the expected reward function $f(x_i)$ using a grid of input values, $x_i \in \{x_0, x_1, \ldots, x_N\}$. This paper focuses on one-dimensional cases. The approximation then takes the following recursive form:  \[f(x_i) = f(x_{i-1}) + \nabla f(x_{i-1}) \cdot (x_i - x_{i-1}),\] for $i \in \{1, \ldots, N\}$.
\begin{figure}[!t]
\centering
\includegraphics[width=2.5in]{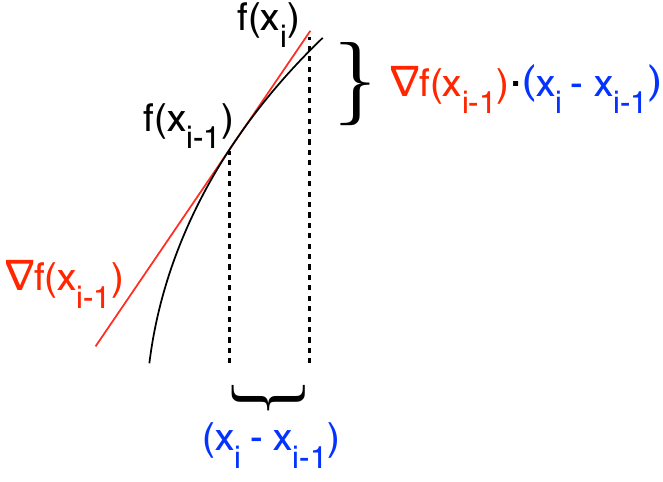}
\caption{Linear approximation of $f(x_i)$ based on the preceding function value $f(x_{i-1})$ and gradient $\nabla f(x_{i-1})$.}
\label{fig_approx}
\end{figure}
As illustrated in Fig. \ref{fig_approx}, for any input value $x_i$, the function value, $f(x_i)$,  is formulated in terms of the gradient, $\nabla f(x_{i-1})$, and the function value, $f(x_{i-1})$, of the preceding point, $x_{i-1}$. Note that for $x_0$, $f(x_0)$ then becomes a constant. In all brevity, the gradient and function values are related in a manner that allows the underlying function to be approximated with arbitrary accuracy.  That is, the approximation can be made arbitrarily accurate by making the grid increasingly fine grained: $\lim_{h\to0}\|f(x + h) - f(x) - \nabla f(x) \cdot h\| = 0$ (for any differentiable function $f$).

We are now ready to present our novel Bayesian scheme that explicitly connects the expected reward function with its gradient.
\begin{figure}[!t]
\centering
\includegraphics[width=3.0in]{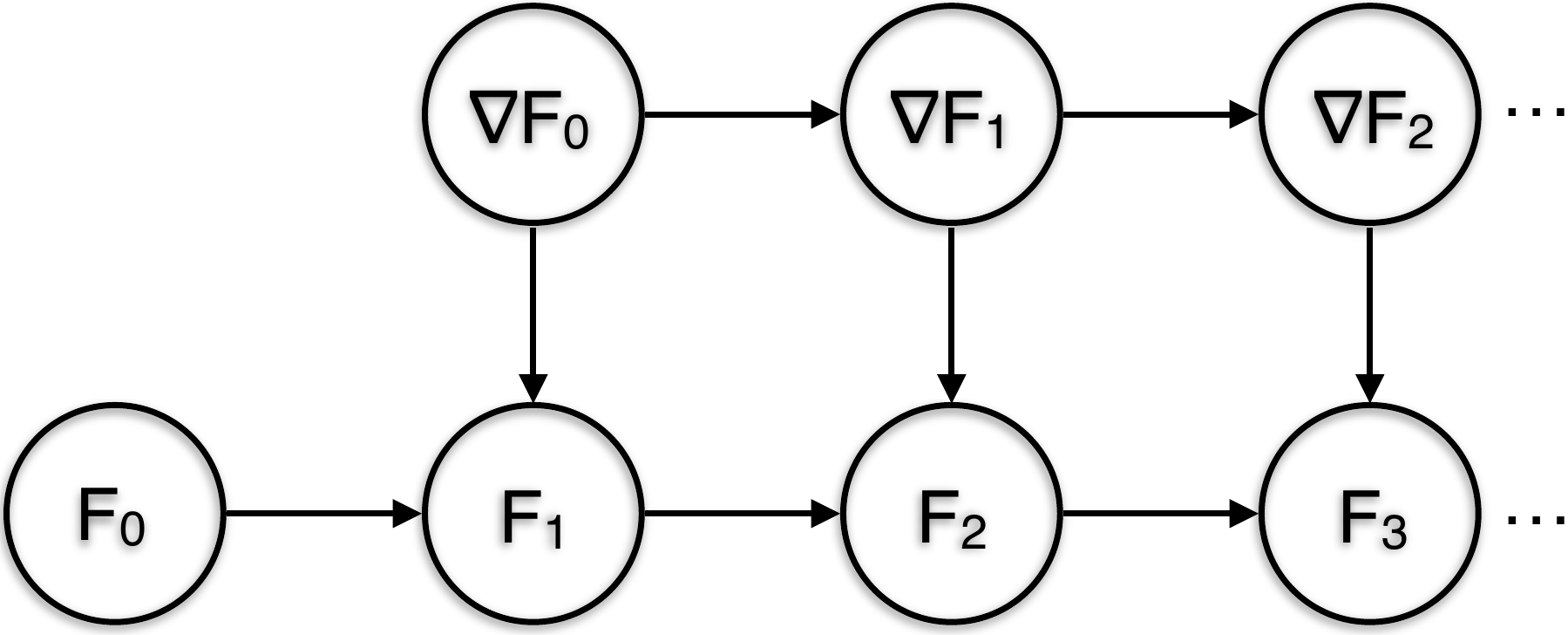}
\caption{Relating gradient and function values for unified gradient and bandit based learning.}
\label{fig_bug-b}
\end{figure}
As shown in Fig. \ref{fig_bug-b}, we model each $f(x_i)$ and $\nabla f(x_i)$  as stochastic variables $F_i$ and $\nabla F_i$, respectively. These stochastic variables are normally distributed, with corresponding unknown means, $\mu_i$ and $\mu_{i\nabla}$, and variations, $\sigma^2_i$ and $\sigma^2_{i\nabla}$. As further seen in the figure, the relationship between variables are defined recursively, according to the aforementioned linear approximation scheme: $F_i = F_{i-1}  + \nabla F_{i-1} (x_i - x_{i-1}) + \epsilon^f_i$. Here, $\epsilon^f_i$ captures uncertainty, representing i.i.d. Gaussian noise  $\epsilon^f_i \sim N(0,\sigma^2_f)$. Furthermore, we model the dynamics of the unknown gradient $\nabla f$ of $f$  by relating neighbouring stochastic gradient variables: $\nabla F_i 
= \nabla F_{i-1} + \epsilon^g_i$. That is, the change rate is stochastically governed by i.i.d. Gaussian noise $\epsilon^g_i \sim N(0, \sigma^2_g)$. 

Using a factor graph based computation approach for the above model, we can efficiently calculate the posterior joint and marginal distributions for all the  variables, given noisy information on both function and gradient values (the computational complexity grows linearly with the number of grid points). 

\begin{figure*}[!ht]
\begin{subfigure}{.5\textwidth}
\centering
\includegraphics[width=3.0in]{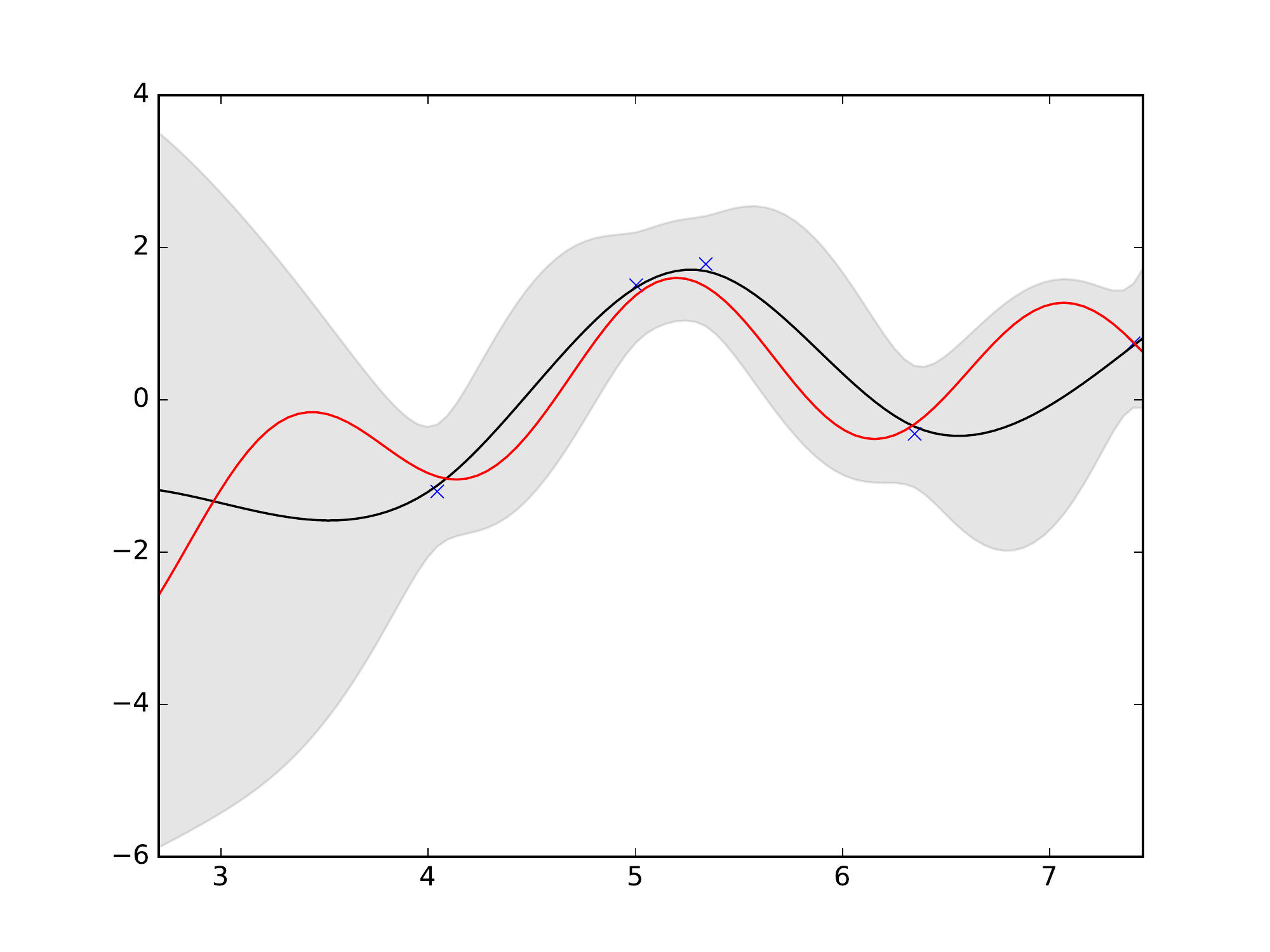}
\caption{BUG-B w/o gradient feedback}
\end{subfigure}
\begin{subfigure}{.5\textwidth}
\centering
\includegraphics[width=3.0in]{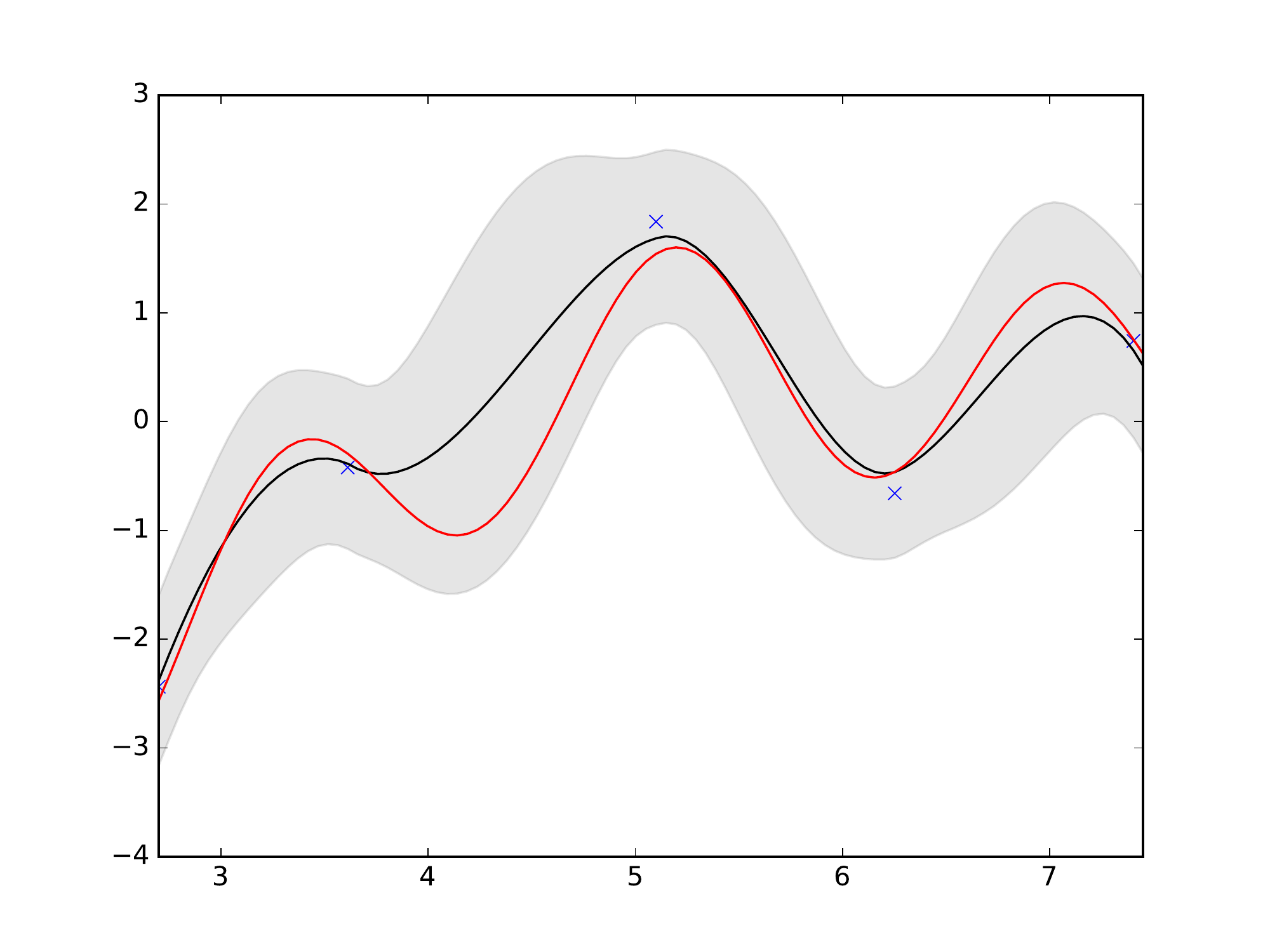}
\caption{BUG-B w/gradient feedback}
\end{subfigure}
\caption{Prediction after five observations with BUG-B, with the observation points selected using a Bayesian upper credible region strategy (red line marks the original functions, while 
the black line marks the prediction along with a 99\% credible interval). }
\label{fig_bug-b_example}
\end{figure*}

\subsection{Optimization Strategies with Thompson Sampling and Upper Confidence Intervals}

From a rather broad perspective, there are currently two competing strategies for finding the global optimum in the bandit setting:  Thompson sampling (stochastic probability matching schemes) and those based on upper confidence (or credibility) bounds. Thompson sampling tends to provide better performance than 
UCB-based approaches in empirical investigations, however, is known to over-explore. UCB-like approaches, on the other hand, provide a deterministic and more goal-directed path towards the global optimum, finding the optimum with probability arbitrarily close to unity. Thompson sampling, on the other hand, \emph{always} converges to the global optimum (with unit probability). \cite{Granmo2010a} 

In \cite{Granmo2010a} we proposed a \emph{Bayesian} technique for
solving bandit like problems, akin to the \textit{Thompson Sampling}
\cite{Thompson1933} principle, leading to novel schemes for handling
multi-armed and non-stationary (restless) bandit problems
\cite{Granmo2010c,Granmo2010b}. Empirical results demonstrated the
advantages of these techniques over established top
performers. Furthermore, we provided theoretical results stating that
the original technique is instantaneously self-correcting and that it
converges to only pulling the optimal arm with probability as close to
unity as desired. Later on, as a further testimony to the renewed importance
of the Thompson Sampling principle, a modern Bayesian look at the
multi-armed bandit problem was also taken in \cite{Scott2010,May2011}.

A promising avenue for solving the multi-armed bandit problem involves the methods which
consider the estimation of  confidence intervals, wherein the scheme
estimates a confidence interval for the reward probability of each
arm, and an ``optimistic reward probability estimate'' is identified
for each arm. The arm with the most optimistic reward probability
estimate is then greedily selected \cite{Vermorel2005,Kaelbling1993}.

In \cite{Auer2002}, the authors analysed several confidence interval
based algorithms. These algorithms also provide logarithmically
increasing \emph{regret}, with \emph{UCB-Tuned} -- a variant of the
well-known \emph{UCB1} algorithm --- outperforming both \emph{UCB1},
\emph{UCB2}, as well as the $\epsilon_n$-greedy strategy. In brief,
in \emph{UCB-Tuned}, the following optimistic estimates are used for
each arm $i$:
\begin{equation}
\mu_i + \sqrt{{\ln n \over n_i} \min\left\{1/4, \sigma_i^2 + \sqrt{{2 \ln n \over n_i}}\right\}}
\end{equation}
where $\mu_i$ and $\sigma_i^2$ are the sample mean and variance of
the rewards that have been obtained from arm $i$, $n$ is the total
number of arm pulls, and $n_i$ is the number of times arm $i$ has
been pulled. Thus, the quantity added to the sample average of a
specific arm $i$ is steadily reduced as the arm is pulled, and the
corresponding uncertainty about the reward probability is reduced.
As a result, by always selecting the arm with the highest optimistic
reward estimate, \emph{UCB-Tuned }gradually shifts from exploration
to exploitation.

\emph{By providing a Bayesian estimate of the function $f(x)$ to be optimised, the BUG-B model supports both Thompson Sampling and UCB-based optimization.} However, as further explored below, we obtained best performance by calculating 95\% Bayesian Credible Bounds across the grid of input values, $x_i \in \{x_0, x_1, \ldots, x_N\}$. By iteratively measuring the function value at the highest bound and then updating our estimate for $f(x)$ using BUG-B, we were able to quickly converge to the maxima of the function.

\section{Empirical Results}
\label{results}

In this section we evaluate the BUG-B scheme by comparing it with
the currently best performing approaches. Based on
our comparison with these ``reference'' algorithms, it should be quite
straightforward to also relate the BUG-B performance results to the
performance of other similar algorithms. 

\subsection{Experimental Setup}

 We have conducted numerous experiments using various functions, generating artificial data, under varying degrees of observation noise. The
 full range of empirical results all show the same trend, however, we
 here report performance on a representative subset of the experiment
 configurations, involving uni-modal and multi-modal functions, with varying degrees of 
 noise and resolutions.
 Performance is measured in terms of \emph{Regret} --- \emph{the
   difference between the sum of rewards expected after $N$ successive
   rounds  and what would have been obtained by always selecting the optimal point}.

For these experiment configurations, an ensemble of $1000$ independent
replications with different random number streams was performed to
minimize the variance of the reported results. In order to investigate
the performance of the schemes under a broad spectrum of environments,
we test the schemes using three different representative 
functions --- one sloped, with a single maxima, and one more peaked with multiple local maxima, particularly similar to the global maxima. To investigate performance 
under varying degrees of noise we introduced i.i.d. Gaussian observation noise, $N(0, \sigma_o)$, employing a diverse range of noise levels: $\sigma_o \in \{0.01, 0.1, 1.0, 5.0\}$.
Regret is reported after 25, 50, 100, and 250 iterations for both
the new \emph{accelerating} scheme and the traditional \emph{static}
scheme.

\subsection{Comparison of Regret}
The \emph{regret} measure is non-trivial, and so we
   provide further clarification here.
In brief, the \emph{regret } can be seen as
 \emph{the difference between the sum of rewards expected after $N$
   successive arm pulls, and what would have been obtained by only
   pulling the optimal arm}.  To exemplify, assume that a
   \emph{reward} yields a
 value (utility) of $1$, and that a \emph{penalty} is associated with the
 value $0$. This implies that the expected utility of pulling arm $i$
 is $r_i$. Thus, if the optimal arm is arm $1$, the \emph{regret} after
 $N$ plays would become:
\begin{equation}
r_1 N - \sum_{n=1}^N \hat{r}_n,
\end{equation}
with $\hat{r}_n$ being the expected reward at arm pull $n$, given
the agent's arm selection strategy.

Table \ref{table_regret_time_steps} reports average regret over multiple functions for different number of time steps. As exemplified in Fig. \ref{fig_bug-b_example}, pursuing a UCB 
strategy, BUG-B only needs 4-5 observations to capture the underlying function, allowing it to quickly zoom in on the most promising input value regions. The effect of this is seen in the 
table, with BUG-B performing significantly better than the competing state-of-the-art schemes.

Also notice how BUG-B outperforms the Gaussian process based UCB approach, even 
when not receiving feedback on the gradient function. This could be explained by the ability of BUG-B to {\it infer}  gradient information indirectly by means of the noisy  function value observations. 
\begin{table}[!t]
\renewcommand{\arraystretch}{1.3}
\caption{Comparison of regret after increasing number of iterations (grid resolution 100, noise 1.0)}
\label{table_regret_time_steps}
\centering
\begin{tabular}{|c||c|c|c|c|}
\hline
{\bf Algorithm / Time steps} & {\bf 25} & {\bf 50} & {\bf 100}&{\bf 250}\\
\hline
BUG-B w/o gradient&20.0&26.4&34.2&50.9\\
BUG-B&{\it 13.3}&{\it 16.7}&{\it 21.8}&{\it 31.5}\\
Multi-armed Bandit w/UCB&26.0&35.9&48.5&71.7 \\
Gaussian Process w/UCB &19.9&27.1&36.3&53.9 \\
Gradient Descent &23.0&43.6&79.2&171.9 \\
Uniform &38.1&76.4&152.8&381.1 \\
\hline
\end{tabular}
\end{table}

\begin{figure}[!t]
\centering
\includegraphics[width=4.5in]{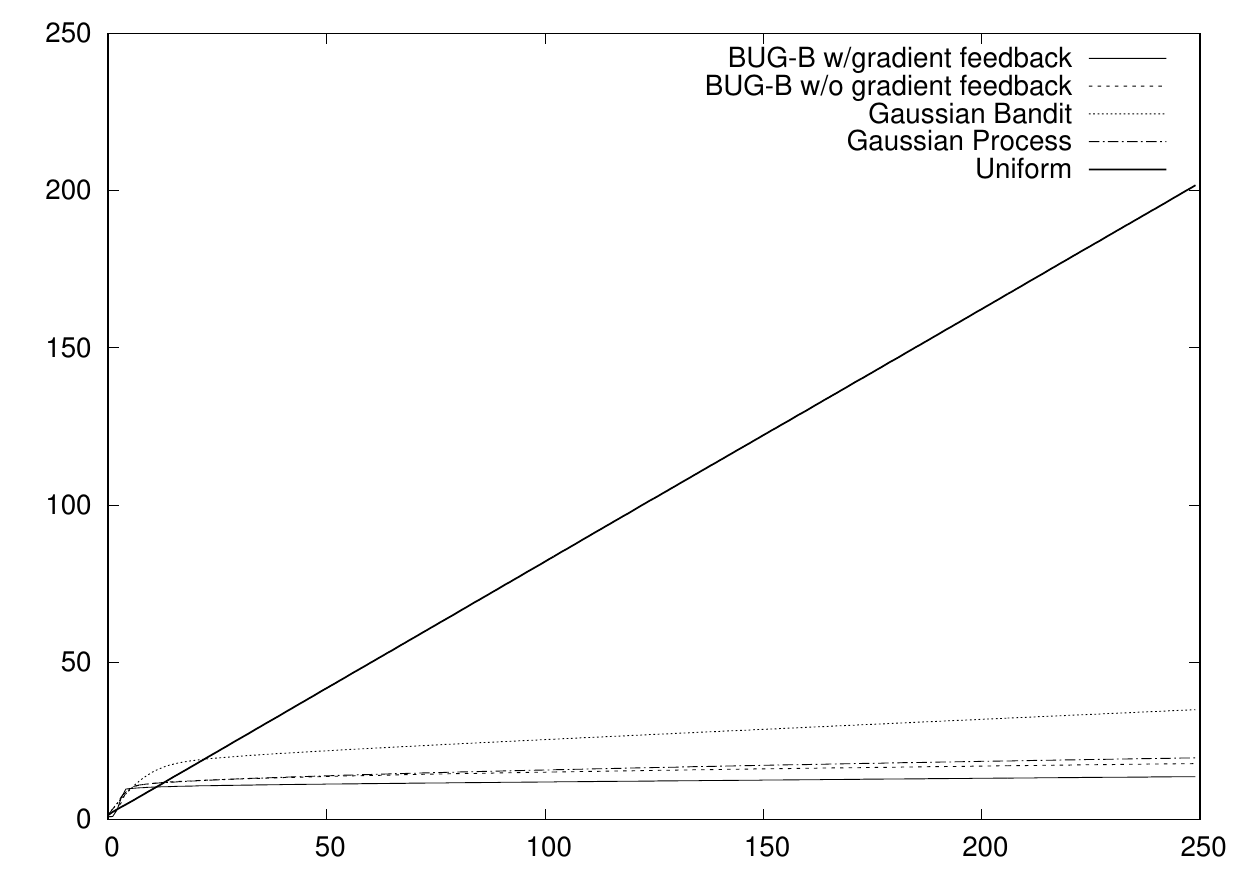}
\caption{Regret plot over time}
\label{fig_regret_plot}
\end{figure}

The above findings are confirmed by the plots in Fig. \ref{fig_regret_plot}, showing that BUG-B provides superior performance at every time step.The Gaussian process based approach is better than BUG-B without gradient feedback up to time step ten or so, and then BUG-B w/o gradients is slightly better for the remainder of the time steps.

Table \ref{table_regret_noise} summarises performance under a diverse range of noise levels, from $0.01$ up to $5.0$. BUG-B is consistently the superior approach  across all the noise 
levels, both with and without feedback on the gradient.
\begin{table}[!t]
%% increase table row spacing, adjust to taste
\renewcommand{\arraystretch}{1.3}
% if using array.sty, it might be a good idea to tweak the value of
% \extrarowheight as needed to properly center the text within the cells
\caption{Comparison of regret under varying degrees of noise  (grid resolution 100)}
\label{table_regret_noise}
\centering
\begin{tabular}{|c||c|c|c|c|}
\hline
{\bf Algorithm / Noise} & {\bf 0.01} & {\bf 0.1} & {\bf 1.0}&{\bf 5.0}\\
\hline
BUG-B w/o Gradient&12.8&17.8&50.9&110.57\\
BUG-B&{\it 11.5}&{\it 13.6}&{\it 31.5}&{\it 55.6}\\
Multi-armed Bandit w/UCB&31.6&34.9&71.7&149.2\\
Gaussian Process w/UCB&14.4&19.6&53.9&175.6 \\
Gradient Descent &212.3&201.7&171.9&249.6 \\
Uniform &382.1&381.7&381.1&380.5 \\
\hline
\end{tabular}
\end{table}
Interestingly,  gradient descent improves performance in the mid range noise levels. This can be explained by the increased noise opening up for escaping local optima, however, performance falls 
again with the largest degree of noise.

Table \ref{table_computation_time} summarises computational performance. In all brevity, the model structure of BUG-G lends itself to efficient computation by exploiting model structure for local computation. This leads to linear 
increase in computation time with respect to number of observations, as opposed to the much more computationally expensive Gaussian process based approach (with exact computation involving covariance matrix inversion).
\begin{table}[!t]
%% increase table row spacing, adjust to taste
\renewcommand{\arraystretch}{1.3}
% if using array.sty, it might be a good idea to tweak the value of
% \extrarowheight as needed to properly center the text within the cells
\caption{Comparison of computation time}
\label{table_computation_time}
\centering
\begin{tabular}{|c||c|c|c|c|}
\hline
{\bf BUG-B} & {\bf MAB} & {\bf Gaussian Process} & {\bf Gradient Descent}\\
\hline
1.74 & 0.58 & 789.7 &{\it 0.04} \\
\hline
\end{tabular}
\end{table}
For all of these experiments, the gradient of the functions were pre-calculated, making gradient descent computationally extremely efficient.

\section{Conclusions and Further Work}
\label{conclusion}

In this paper we have proposed a novel approach to global optimisation where bandit based and gradient based learning is combined. Our Bayesian model, BUG-B, unifies the two paradigms in one integrated model. At the heart of the model we find a stochastic linear approximation of the function to be optimised. Here, both the gradient and function values are explicitly related.  This allows us to learn from both noisy function and gradient observations, as well as predicting these properties across the action space to support optimisation.

We further proposed an accompanying bandit driven exploration scheme that use Bayesian credibility bounds to trade off exploration against exploitation.  Our empirical results demonstrated that by unifying bandit and gradient based learning, one obtains consistently improved performance across a wide spectrum of environments. Furthermore, even when gradient feedback is unavailable, the flexibility of our model, including gradient prediction, allows us to still outperform competing approaches, although with a smaller margin. Due to the pervasiveness of bandit based optimisation, our scheme  opens up for improved performance both in meta-optimisation and in applications where gradient information is readily available.

In future work, we propose that these pioneering results are expanded in a number of directions. First of all, BUG-B needs to be generalised to cover multi-dimensional functions. Additionally, formal regret bounds and asymptotic properties needs to be established. Finally, it would be interesting to investigate how BUG-B can be leveraged in novel application areas, such as meta-learning in neural networks.

\bibliographystyle{IEEEtran}
\bibliography{Granmo}

\end{document}